\documentclass[default]{sn-jnl}

\usepackage{graphicx}%
\usepackage{multirow}%
\usepackage{amsmath,amssymb,amsfonts}%
\usepackage{amsthm}%
\usepackage{mathrsfs}%
\usepackage[title]{appendix}%
\usepackage{textcomp}%
\usepackage{manyfoot}%
\usepackage{booktabs}%
\usepackage{listings}%

\usepackage{url}
\usepackage{hyperref}
\usepackage{subcaption}
\usepackage{graphicx}
\usepackage{amsmath,amssymb,amsfonts}
\usepackage{physics}  
\usepackage{natbib}
\usepackage[linesnumbered, vlined, algo2e, ruled, noend]{algorithm2e}
\usepackage{xcolor}

\theoremstyle{thmstyleone}%
\theoremstyle{thmstyletwo}%

\theoremstyle{thmstylethree}%

\DeclareMathOperator{\NN}{\mathrm{NN}}
\DeclareMathOperator{\ED}{ED}
\DeclareMathOperator{\DTW}{DTW}
\DeclareMathOperator{\DDTW}{DDTW}
\DeclareMathOperator{\WDTW}{WDTW}
\DeclareMathOperator{\WDDTW}{WDDTW}
\DeclareMathOperator{\LCSS}{LCSS}
\DeclareMathOperator{\ERP}{ERP}
\DeclareMathOperator{\MSM}{MSM}
\DeclareMathOperator{\TWE}{TWE}
\DeclareMathOperator{\cDTW}{cDTW}
\DeclareMathOperator{\cDDTW}{cDDTW}
\DeclareMathOperator{\ADTW}{ADTW}

\def\cwc#1{\textcolor{red}{{\tiny CW:} \bf #1}}

\begin{document}

\title[Proximity Forest 2.0]{Proximity Forest 2.0: A new effective and scalable similarity-based classifier for time series}


\author[1]{\fnm{Matthieu} \sur{Herrmann}}\email{matthieu.herrmann@monash.edu}
\equalcont{These authors contributed equally to this work.}

\author*[1]{\fnm{Chang Wei} \sur{Tan}}\email{chang.tan@monash.edu}

\author[1]{\fnm{Mahsa} \sur{Salehi}}\email{mahsa.salehi@monash.edu}
\equalcont{These authors contributed equally to this work.}

\author[1]{\fnm{Geoffrey I.} \sur{Webb}}\email{geoff.webb@monash.edu}
\equalcont{These authors contributed equally to this work.}

\affil*[1]{\orgdiv{Department of Data Science and AI}, \orgname{Faculty of Information Technology, Monash University, Clayton Campus}, \orgaddress{\street{Woodside Building, 20 Exhibition Walk}, \city{Melbourne}, \postcode{3800}, \state{VIC}, \country{Australia}}}

\abstract{
Time series classification (TSC) is a challenging task due to the diversity of types of features that may be relevant for different classification tasks, including trends, variance, frequency, magnitude, and various patterns. 
To address this challenge, several alternative classes of approach have been developed, including similarity-based, features and intervals, shapelets, dictionary, kernel, neural network, and hybrid approaches. 
While kernel, neural network, and hybrid approaches perform well overall, some specialized approaches are better suited for specific tasks. 

In this paper, we propose a new similarity-based classifier, Proximity Forest version 2.0 (PF 2.0), which outperforms  previous state-of-the-art similarity-based classifiers across the UCR benchmark and outperforms state-of-the-art kernel, neural network, and hybrid methods on specific datasets in the benchmark that are best addressed by similarity-base methods. 
PF 2.0 incorporates three recent advances in time series similarity measures --- (1) computationally efficient early abandoning and pruning to speedup elastic similarity computations; (2) a new elastic similarity measure, Amerced Dynamic Time Warping ($\ADTW$); and (3) cost function tuning. It rationalizes the set of similarity measures employed, reducing the eight base measures of the original PF to three and using the first derivative transform with all similarity measures, rather than a limited subset. 
We have implemented both PF 1.0 and PF 2.0 in a single C++ framework, making the PF framework more efficient.
}

\maketitle

\section{Introduction}
\label{sec:intro}
One of the challenging features of time series classification (TSC) is that there is extraordinary diversity in the forms of feature that may be relevant to classification, including trends, variance, frequency, magnitude, first, second or further derivatives, local patterns and global patterns. 
This diversity in what aspect of a series might be relevant to a given classification task has led to the development of a plethora of alternative approaches. 
These include 
\emph{similarity based approaches} \citep{lucas2019proximity,lines2015time,tan2020fastee,ding2008querying},
\emph{features and intervals approaches} \citep{fulcher2017hctsa,lubba2019catch22,middlehurst2020canonical,middlehurst2021temporal}, 
\emph{shapelets} \citep{bostrom2017binary,middlehurst2021hive}, 
\emph{dictionary approaches} \citep{large2019time,middlehurst2021temporal,dempster2022hydra}, 
\emph{kernel approaches} \citep{dempster2020rocket,dempster2021minirocket,tan2022multirocket}, 
\emph{neural networks} \citep{wang2017time,ismail2020inceptiontime} and 
\emph{hybrid approaches} \citep{middlehurst2021hive,shifaz2020ts}.

\begin{figure}[t]
    \centering
    \begin{subfigure}{0.49\textwidth}
        \includegraphics[width=\textwidth]{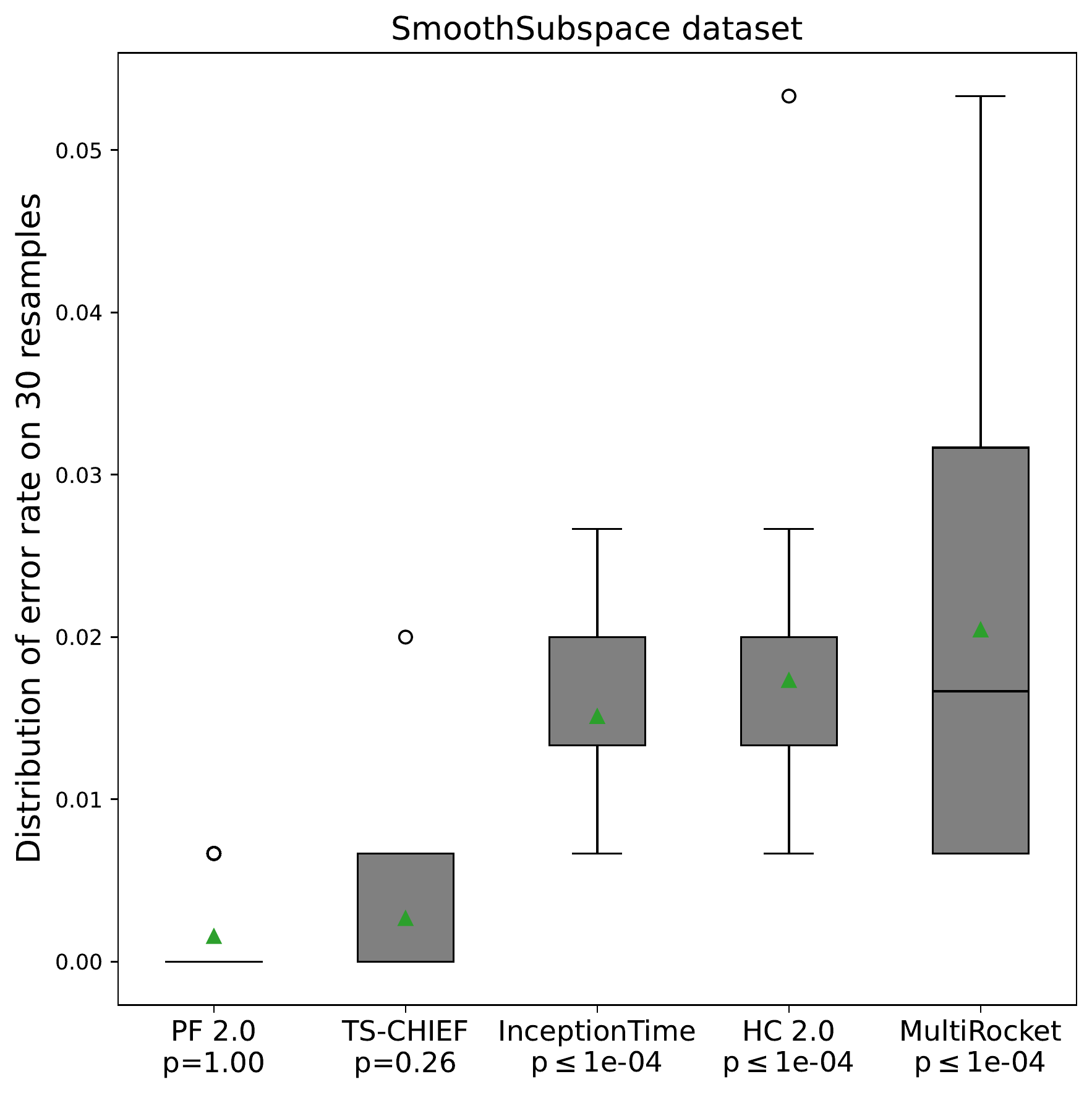}
        \caption{}
        \label{fig:bar smoothsubspace}
    \end{subfigure}
    \begin{subfigure}{0.49\textwidth}
        \includegraphics[width=\textwidth]{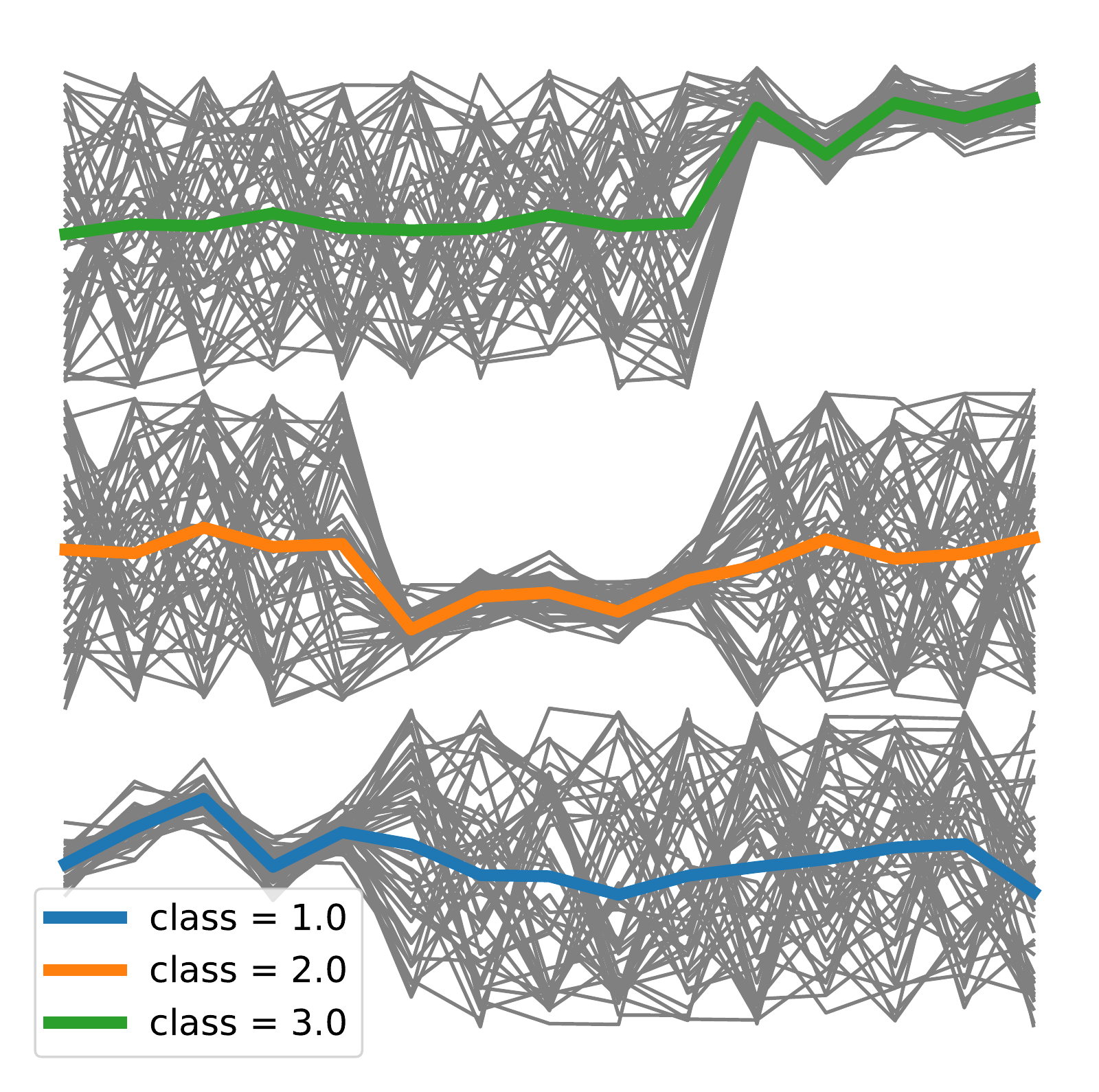}
        \caption{}
        \label{fig:ts smoothsubspace}
    \end{subfigure}
    
    \caption{(a) Boxplots for comparing the error rate of PF 2.0 with the top 4 state-of-the-art (SoTA) TSC algorithms on the \texttt{SmoothSubspace} dataset taken from the UCR time series archive over 30 resamples. The p-values are computed between PF 2.0 and SoTA using a t-test. The triangles represent the mean error rate. (b) Time series in the \texttt{SmoothSubspace} dataset with the average series of each class. The discriminatory pattern of class 1 is at the start, class 2 in the middle and class 3 is at the end.}
    \label{fig:smoothsubspace}
\end{figure}

In benchmark evaluation on the UCR repository \citep{UCRArchive2018}, kernel, neural network and hybrid approaches dominate in terms of overall performance. 
However, there remain specific tasks for which each of the more specialized approaches dominate \citep{bagnall2017great}. 
One example is the SmoothSubspace dataset from the UCR repository, for which similarity based approaches dominate, as illustrated in Figure~\ref{fig:smoothsubspace}, which shows the error on this dataset of our proposed new similarity-based classifier Proximity Forest version 2.0 (PF 2.0) relative to the leading kernel (MultiRocket \citep{tan2022multirocket}), neural network (InceptionTime \citep{ismail2020inceptiontime}) and hybrid (HIVE COTE 2 \citep{middlehurst2021hive} and TS-Chief \citep{shifaz2020ts}) methods.
Our new Proximity Forest (PF) 2.0 achieves significantly lower error than any of these four state of the art methods. 

The first Proximity Forest (which we refer to in the paper as PF 1.0) has been the best performing similarity-based classifier on the UCR benchmark since it was introduced \citep{lucas2019proximity}.
PF 2.0 makes multiple fundamental changes to PF 1.0. 
It incorporates three important recent advances in time series similarity measures -- 
\begin{enumerate}
\item computationally efficient early abandoning and pruning to speedup elastic similarity computations \citep{herrmann2021early}; 
\item a new similarity measure Amerced Dynamic Time Warping ($\ADTW$) \citep{herrmann2023amercing}; and 
\item cost function tuning \citep{herrmann2023parameterizing}. 
\end{enumerate}
PF 2.0 also rationalizes the set of similarity measures that are employed.
PF~1.0 directly emulated the similarity measures employed by the Elastic Ensemble \citep{lines2015time} in order to allow direct comparability. 
PF 2.0 introduces the new $\ADTW$ and removes all the remaining similarity measures other than Dynamic Time Warping with windowing ($\cDTW$) and Longest Common Sub Sequence ($\LCSS$). 
Finally, it allows the first derivative transform to be employed with all similarity measures, while PF 1.0 only paired it with some variants of $\DTW$. We also demonstrate that this rationalized set of similarity measures has potential application beyond Proximity Forest, showing that it supports a fast and accurate variant of the Elastic Ensemble. 

PF 2.0 is significantly faster and more accurate on the UCR benchmark than its predecessor.
We have implemented both PF 1.0 and PF 2.0 in a single C++ framework, making PF 1.0 more efficient than the previous Java implementation and supporting fair comparison of computational performance.
Our paper is organised as follows. 
Section \ref{sec:background and related work} describes key background and related work. 
We illustrate and describe our new PF 2.0 in detail in Section~\ref{sec:pf2}.
After that we provide and discuss comprehensive experimental results in Section~\ref{sec:experiments}.
Finally we conclude our work in Section~\ref{sec:conclusion}.

\section{Background and Related Work}
\label{sec:background and related work}
This section provides a brief background overview of time series classification and key related work.
We refer interested readers to the paper \citep{bagnall2017great} for a more detailed survey and discussion on TSC.






A time series is a sequence of data points measuring a phenomenon over time. 
In addition to signals ordered in time (such as a sensor signal), time series techniques are often applied to other numeric series such as spectra. The~UCR time series benchmark includes a number of such series.  
TSC learns a function from time series $S=\{s_1,s_2,...s_L\}$ of length $L$ to a discrete class label $c$ \citep{bagnall2017great}. 
We will focus our attention on univariate time series, where $s_i\in S$ is a 1-dimensional point, with equal length, as this is where most of the research in TSC has been focused.
We note that there is also work on multivariate time series \citep{ruiz2021great,shifaz2023elastic}, where $s_i\in S$ is a $d$-dimensional point, and time series of variable length \citep{tan2019time} and will consider these as future work. 

TSC algorithms are usually categorised by the core data representation used. 
Similarity-based algorithms (Section \ref{sec:similarity-based}) work on the raw time series, i.e. time series that are not transformed to any other domains or representations \citep{lines2015time,tan2020fastee,ding2008querying}.
They rely on elastic similarity measures to match the shape and calculates the similarity between two series.

Shapelets-based approaches are similar to similarity-based approaches.
They calculate the similarity between a series and a phase discriminatory subsequence called shapelets.
One of the most accurate shapelets-based approaches -- the Shapelets Transform (ST) algorithm transforms a time series using its similarity to all shapelets and trains a Rotation Forest for classification \citep{bostrom2017binary,middlehurst2021hive}. 
Instead of finding a single phase discriminatory pattern in the series, dictionary-based methods such as HYDRA \citep{dempster2022hydra} and Temporal Dictionary Ensemble (TDE) \citep{middlehurst2021temporal} transform the series into `words', which are commonly interpreted as \emph{recurring patterns} in the series. 
They work by comparing the frequency of each word between two series. 

Features and interval-based algorithms extract features from intervals of the series and train a machine learning (ML) algorithm for the classification task.
They are more effective than algorithms that extract features from the whole series due to their ability to find relevant temporal features.
The most accurate interval-based algorithm, DrCIF is a forest of time series trees. 
It~works with multiple representations and extracts random set of features from random intervals per tree \citep{middlehurst2020canonical,middlehurst2021temporal}. 

The majority of traditional TSC algorithms only work with a single representation. 
However, a number of more recent TSC algorithms work with multiple representations.
They exploit the fact that different problems require different representations and leverage on the relationship between the different representations to achieve high classification performance.
Convolution and transformation approaches, such as the Rocket approaches \citep{dempster2020rocket,dempster2021minirocket,tan2022multirocket} and MrSQM \citep{le2019interpretable}, have become popular due to their scalability and accuracy. 
They map the time series into a high-dimensional space by creating massive features using large number of convolutions and transformation operations.
This massive feature space captures multiple representations of the series, allowing them to achieve superior classification accuracy.
The most accurate TSC algorithm HIVE-COTE 2.0 is a meta-ensemble that consists of 4 main ensemble classifiers, each of them being one of the best classifiers in their respective representation.
TS-CHIEF \citep{shifaz2020ts} is a tree-based homogeneous ensemble where the different representations are embedded within the nodes of the tree.
Deep learning algorithms learn the best representations for the time series problem through a neural network \citep{ismail2019deep}. 
Deep learning approaches such as InceptionTime \citep{ismail2020inceptiontime}, are powerful because they have the capabilities to learn latent representations of the series that generalise well to new time series.

\subsection{Similarity-based time series classification}
\label{sec:similarity-based}

\subsubsection{Time series similarity measures}
\label{sec:time series similarity}
Time series are often auto-correlated, where the value of the time series at a timestamp is likely to be close to the ones immediately before and after. 
There can also be non-linear distortions in the time axis caused by the different start and end times or frequency of the observed phenomenon.
These factors require specific similarity measures that take into account auto-correlated values and non-linear distortions when calculating the similarity of a pair of time series data. 

There have been many similarity measures developed for time series data. 
The majority of them have at least one hyper-parameter, allowing them to be tuned to a specific time series problem and increasing their accuracy.

The simplest and fastest measures to compute are the Euclidean and its generalisation, the Minkowski distance \citep{paparrizos2020debunking}.
Although they do not consider the non-linear distortions in the time axis, they have proven to be effective for many time series datasets \citep{paparrizos2020debunking}, especially when the training dataset is large \citep{ding2008querying}.

Dynamic Time Warping ($\DTW$) \citep{sakoe1972dynamic} is a popular similarity measure for comparing time series and used in many applications. 
Many similarity measures have been developed around $\DTW$, such as the constrained $\DTW$ ($\cDTW$) -- $\DTW$ with a learnt warping window;
Derivative $\DTW$ \citep{keogh2001derivative} transforms the time series into their first order derivatives before applying $\DTW$;
Weighted $\DTW$ ($\WDTW$) \citep{jeong2011weighted} and the recent Amerced $\DTW$ ($\ADTW$) \citep{herrmann2023amercing} that apply weights to off-diagnonal $\DTW$ alignments. 

Some other popular measures include the Longest Common Subsequences ($\LCSS$) that was modified from string matching \citep{vlachos2006indexing}, Edit distance with Real Penalty ($\ERP$) \citep{chen2005robust,chen2004marriage}, Time Warp Edit Distance ($\TWE$) that allows the alignment of timestamps \citep{marteau2008time} and Move Split Merge ($\MSM$) \citep{stefan2012move}.

Most of these measures, other than the recent $\ADTW$, have been employed by the Ensemble of Elastic Distances (EE) \citep{lines2015time,bagnall2017great} and the first Proximity Forest (PF 1.0) \citep{lucas2019proximity}.
Such ensembles deliver substantial improvements in accuracy relative to $\NN$ classifiers using any single measure.

We will describe $\ADTW$ in more detail (in Section \ref{sec:adtw}) and refer readers to the mentioned works for a more complete description of other measures.

\subsubsection{Nearest neighbour classification}
\label{sec:nn}
The most common similarity-based TSC method is pairing each similarity measure with the nearest neighbour ($\NN$) algorithm \citep{ding2008querying,rakthanmanon2012searching,ratanamahatana2004making,tan2017indexing,tan2018efficient,bagnall2017great}.
Given a similarity measure, the $\NN$ algorithm classifies a query time series based on its similarity (proximity) to another time series in a training dataset.
The $\NN{-}\DTW$ with its warping window (parameter) tuned was a strong TSC baseline for more than a decade \citep{lines2015time}.
Many measures with different properties and alignment strategies were proposed \citep{jeong2011weighted,stefan2012move,vlachos2006indexing,marteau2008time,chen2005robust} but none of them were able to significantly outperform $\DTW$.
The Ensemble of Elastic Distances (EE) was the first algorithm that was significantly more accurate than $\NN{-}\DTW$ \citep{lines2015time}.
It ensembles eleven popular similarity measures and weights each similarity measure based on their training accuracy (each similarity measure is paired with a $\NN$ classifier and the parameters are fine-tuned using leave-one-out cross validation).

Despite being more accurate than any individual $\NN$ classifier, EE is computationally expensive. 
Training and tuning the parameters of a typical $\NN$ classifier in EE takes $O(N^2L^3)$ operations \citep{tan2018efficient,tan2020fastee}.
This has led to research into speeding up $\NN$ classifiers \cite{tan2017indexing,keogh2005exact,tan2019elastic,webb_lowerbounds}, similarity computations \citep{silva2018speeding,herrmann2021early} and the training process of $\NN$ \citep{tan2018efficient,tan2021ultra,tan2023ultra}.
One of the first attempts used computationally efficient lower bounds to prune unpromising candidates, avoiding the expensive similarity computation \citep{keogh2005exact,rakthanmanon2012searching,tan2019elastic}.
FastEE \citep{tan2020fastee} is a faster version of EE that employs lower bounds for all the measures used in EE and an efficient hyper-parameter search framework for time series $\NN$ classifiers \citep{tan2018efficient}.
\citet{lines2022ts} recently proposed TS-QUAD, a smaller version of EE that only uses 4 similarity measures, $\WDTW$, $\DDTW$, $\MSM$ and $\LCSS$.  
They showed that TS-QUAD (EE with only 4 measures) was able to match the performance of the original EE, while at least halving the run-time.

Another line of research has investigated how to efficiently compute the $O(L^2)$ time series similarity measures.
The first attempt was to early abandon $\DTW$ and Euclidean distance computation as soon as an abandoning criteria is met \citep{rakthanmanon2012searching}.
This is achieved by monitoring the minimal cost of the measure at any point of the computed paths, and abandoning when it exceeds a cut-off \citep{rakthanmanon2012searching}. 
For nearest neighbor search, the cut-off is the similarity to the nearest neighbour so far.
The PrunedDTW algorithm was later proposed to `prune' $\DTW$ paths that must exceed the threshold  \citep{silva2018speeding}.
The recently proposed early abandoning and pruning algorithm (EAP) tightly integrates both early abandoning and pruning strategies from previous work to efficiently compute 6 time series similarity measures including $\DTW$ \citep{herrmann2021early}.
The EAP algorithm demonstrated more than an order of magnitude speedup for several $\NN{-}\DTW$ search tasks \citep{herrmann2021early}.

A $\NN$ classifier paired with a time series simiilarity measure is typically tuned using leave-one-out cross validation (LOOCV), which is a time consuming process. 
\citet{tan2018efficient} was the first to study and exploit the properties of $\DTW$ to speed up LOOCV for time series $\NN$ classifier.
They proposed the fast parameter search framework that was later extended to other measures in FastEE \citep{tan2020fastee}.
The method is exact with up to 3 magnitudes of speed up compared to LOOCV. 
The fast parameter search framework relies on good lower bounds to speed up the cross validation process.
The development of EAP implementations of time series similarity measures extends the work in \citep{tan2018efficient} to ultra fast parameter search framework \citep{tan2021ultra,tan2023ultra}. 
The ultra fast parameter search framework leverages and exploits the properties of EAP which renders lower bounds redundant.
It is one order of magnitude faster than its predecessor. 

\subsubsection{Proximity Forest}
\label{sec:pf}
Proximity Forest (PF 1.0) \citep{lucas2019proximity} is a forest of tree classifiers called Proximity Trees.
A~Proximity Tree is similar to a regular decision tree, but differs in the tests applied at internal nodes. 
A~Proximity Forest has two main parameters, the number of Proximity Trees, $K$ and the number of candidate splitters, $R$.

A proximity tree differs from a regular decision tree based on the splitting criteria used at each internal node.  
A conventional decision tree splits the data at a node using a threshold on the value of an attribute. In contrast, a proximity tree splits the data based on the \emph{proximity} of each instance to each of a set of  class exemplars, $E$, based on a parameterised similarity measure $\delta$. 
This is referred to as a splitter, $r=(\delta, E)$. 
 
At each node, a set of candidate splitters are selected at random and assessed on the training data that reaches the node. For each splitter, $E$ contains one exemplar per class chosen at random from the training examples that reach the node. Each similarity measure $\delta$ is chosen at random from a set of 11 candidates, each with their own parameter space from which a parameterization is also chosen at random.
Then the candidate splitter with the highest Gini score is selected for that node.
By default, PF 1.0 ensembles 100 Proximity Trees with 5 candidates at each node.

The eleven similarity measures from which $\delta$ is sampled from are Euclidean distance, $\DTW$, $\cDTW$, $\DDTW$, $\cDDTW$, $\WDTW$, $\WDDTW$, $\ERP$, $\MSM$, $\TWE$ and $\LCSS$.  
These were deliberately selected as being the measures used by EE. 
Note that three of these measures, $\DDTW$, $\cDDTW$ and $\WDDTW$ can be considered a combination of a core measure ($\DTW$, $\cDTW$ and $\WDTW$) with the first derivative transform. 
That is, they involve first transforming the series and then computing the similarity with respect to those transforms.
Hence resulting to a total of eight core similarity measures. 


PF 1.0 is more accurate and scalable than EE, and is currently the most accurate similarity-based method on the UCR time series benchmark \citep{UCRArchive2018}.

With the new recent advances in similarity-based methods that will be discussed in the later sections, we believe that it is time to update PF 1.0 and develop a stronger state-of-the-art for similarity-based methods.
We also perform a study on the set of similarity measures used in PF 1.0 to reassess the performance and importance of each measure in the forest.


\subsection{Recent advances in similarity-based classification}
\label{sec:recent advancements} 
There have been many recent advances in similarity-based classification. 
Most of these have focused on speeding up the similarity computation. 
In this work, we focus on two major advances that improve classification accuracy for similarity-based algorithms, 
(1) the development of a new similarity measure, Amerced Dynamic Time Warping; and
(2) the parameterisation of cost functions used in similarity measures. 

\subsubsection{Amerced Dynamic Time Warping}
\label{sec:adtw}
DTW with LOOCV tuning of the window parameter ($\cDTW$) when paired with the $\NN$ algorithm is one of the most accurate similarity-based TSC methods.
It~uses a step function to constrain the alignments, where any warping
is allowed within the warping window, $w$ and none beyond it. 
Although accurate, it is unintuitive for many applications, where some flexibility in the exact amount of warping might be desired.
Recently, the Amerced Dynamic Time Warping ($\ADTW$) similarity measure was proposed  as an intuitive and effective variant of $\DTW$ that applies a tunable additive penalty $\omega$ for non-diagonal (warping) alignments \citep{herrmann2023amercing}.
Similar to $\DTW$ and many other similarity measures, $\ADTW$ is computed with dynamic programming, using a cost matrix $M$ with $\ADTW_{\omega}(S, T) = M(L, L)$.
Equation \ref{eq:ADTW} describes this cost matrix, where $\lambda(s_i, t_j)$ is the cost of aligning the two points.
Section \ref{sec:cost function} discusses this cost function in more detail. $\ADTW$ is identical to $\DTW$ other than the additional terms that add $\omega$ for off-diagonal alignments, colored \textcolor{red}{red}, below.

\begin{subequations}\label{eq:ADTW}
\begin{align}
    M(0,0) &= 0       \label{eq:ADTW:corner}\\
    M(i,0) &= +\infty \label{eq:ADTW:vborder}\\
    M(0,j) &= +\infty \label{eq:ADTW:hborder}\\
    M(i,j) &= \min\left\{
    \begin{aligned}
        &M(i{-}1, j{-}1) + \lambda(s_i, t_j)\\
        &M(i{-}1, j) + \lambda(s_i, t_j) \mathbin{\textcolor{red}{+}} \textcolor{red}{\omega} \\
        &M(i, j{-}1) + \lambda(s_i, t_j) \mathbin{\textcolor{red}{+}} \textcolor{red}{\omega}
    \end{aligned}
    \right. \label{eq:ADTW:main}
\end{align}
\end{subequations}

The parameter $\omega$ serves a similar role to the warping window in $\cDTW$.
It is an additive penalty that is added to off-diagonal alignments, allowing $\ADTW$ to be as flexible as unconstrained $\DTW$ ($\cDTW$ with $w\geq L-2$), or as constrained as Euclidean distance ($\cDTW$ with $w=0$).
A small penalty encourages warping, while a large penalty minimizes warping.

Since $\omega$ is an additive penalty, its scale relative to the time series in context matters.
A value of $\omega$ that is a small penalty in a given problem may be a huge penalty in another one.
In consequence, an automated $\omega$ selection method has been proposed for time series classification that considers the scale of $\omega$ relative to the time series dataset in context \citep{herrmann2023amercing}.
Specifically, the scale of penalties is determined by multiplying the maximum penalty $\omega'$ by a ratio $0 \leq r \leq 1$, i.e. $\omega=\omega'\times r$.
The maximum penalty $\omega'$ is set to the average ``direct alignment'' (diagonal of the cost matrix) of pairs of series sampled randomly from the training dataset.
Then ratios are sampled from $r_i=(\frac{i}{100})^5$ for $1\leq i\leq100$ to form the search space for $\omega$.

$\ADTW$ when used in a $\NN$ classifier is significantly more accurate than $\cDTW$ on the 112 UCR time series benchmark datasets \citep{herrmann2023amercing}.

Note that $\omega$ can be considered as a direct penalty on path length. 
If series $S$ and $T$ have length $L$ and the length of the warping path for $\ADTW_\omega(S,T)$ is $P$, the sum of the $\omega$ terms added will equal $2\omega(P-L+1)$. 
The longer the path, the greater the penalty added by $\omega$. This contrasts to Weighted $\DTW$ ($\WDTW$) \citep{jeong2011weighted}, which applies a multiplicative penalty that increases as distance to the diagonal increases.

\subsubsection{Parameterizing cost functions}
\label{sec:cost function}
Time series similarity measures typically align the points in two series and return the sum of the pairwise distances between each of the pairs of points in the alignment.
Pairwise-distances between two points $(s_i,t_j)$ in series $S$ and $T$ are usually calculated using a cost function $\lambda(s_i,t_j)$.
The common cost functions used in the literature are
(1) \emph{absolute difference}, $\lambda(s_i,t_j)=\|s_i-t_j\|$ and 
(2) \emph{squared difference}, $\lambda(s_i,t_j)=(s_i-t_j)^2$.
The motivations for an exact choice of cost function has not been well articulated in many cases.
The absolute difference cost function was the original cost function when $\DTW$ was introduced \citep{sakoe1972dynamic}.
Within the time series classification community, this has largely been replaced by the squared difference cost function \citep{tan2018efficient,dau2018optimizing,keogh2005exact}.
However, there has been little research into the effects of different types of cost function.

One exploration of different types of cost functions used the Minkowski distance \citep{thompson1996minkowski}.
The Minkowski distance is a generalised form of both the Euclidean and Manhattan distance.  
The Minkowski distance of order $\gamma$ between a univariate time series (vector) of length $L$ is defined in Equation \ref{eq:minkowski}
\begin{equation}
    D^\gamma_{\text{minkowski}}(S,T) = (\sum_{i=1}^{L}{(\|s_i-t_i\|)^\gamma})^{\frac{1}{\gamma}}
    \label{eq:minkowski}
\end{equation}
Common values for $\gamma$ are $\gamma=1$ and $\gamma=2$ that correspond to the Manhattan distance (absolute difference) and the Euclidean distance (squared difference), respectively.
\citet{paparrizos2020debunking} shows that for $\NN$ classification on the UCR benchmark, the Minkowski distance with tuned $\gamma$ outperforms all the other lock-step similarity measures (similarity measures that do not allow non-linear alignments), including the commonly used Euclidean distance.

\citet{herrmann2023parameterizing} explored the effect of tuning a Minkowski cost function for two $\DTW$-based distances, $\cDTW$ and $\ADTW$, by learning the parameter $\gamma$ for a cost function of form $\lambda_\gamma(a,b)=\|a-b\|^\gamma$.  
They showed that tuning the cost function for both $\cDTW$ and $\ADTW$ significantly outperforms their default counterparts where the cost function was not tuned. 
\cite{herrmann2023parameterizing} found that the set $\gamma\in\Gamma=\{1/2, 1/1.5, 1, 1.5, 2\}$ as a good starting point for $\NN$ classifiers on the UCR archive.
Larger and denser sets did not significantly improve the accuracy but increased training time.
Furthermore, the authors also introduced a new variant of PF 1.0, PF$^+$ where the set of cost functions are added as an additional parameter to $\DTW$-based similarity measures ($\DTW$, $\cDTW$, $\WDTW$, $\DDTW$, $\cDDTW$, $\WDDTW$ and $\ED$) in the original PF 1.0 \citep{lucas2019proximity}.
They showed that PF$^+$ is significantly more accurate than PF 1.0.

\section{Proximity Forest 2.0}
\label{sec:pf2}
The new Proximity Forest, Proximity Forest 2.0 (PF 2.0) builds on the structure of the original PF, incorporating the recent advances in similarity-based time series classification outlined in Section \ref{sec:recent advancements}. 
It leverages $\ADTW$ and tuning the cost functions for TSC to increase accuracy, together with the computational efficiency of computing these distances provided by EAP \citep{herrmann2021early}.

PF 2.0 modifies the splitters to comprise three elements, $r=(\delta, \tau, E)$,  a set of class exemplars $E$, a parameterised similarity measure $\delta$ and a time series transform $\tau$.
At each node, each time series is first transformed based on the selected time series transform, $\tau$. The transform is either \emph{raw} (the series is not modified) or \emph{first derivative} (the series is replaced by its first derivative).
Then the similarities between the transformed time series and the exemplars (transformed exemplars if transform is used) are calculated according to the parameterised similarity measure $\delta$.
The time series then follows down the branch corresponding to the exemplar to which it is closest, until it reaches a leaf.
$R$ candidates are uniformly sampled at each node and evaluated.
Then the candidate splitter with the highest Gini score is selected for that node.
We~follow the original PF 1.0 configuration with $K=100$ Proximity Trees with $R=5$ candidates at each node.

Note that the original PF 1.0 also used the first derivative, but following the example of EE, it was only applied in conjunction with some variants of $\DTW$ and its use in this manner was treated as a separate similarity measure.

Algorithm \ref{alg:build tree} presents the algorithm for learning a single proximity tree in PF 2.0. 
The inputs to the algorithm are the labelled time series dataset $D$, a set of parameterized similarity measures $\Delta$, a set of time series transforms $\mathrm{T}$ and the number of candidate splits at each node $R$.
The nodes of the tree are built recursively from the root node down to the leaves.
A node becomes a leaf when the node is pure, i.e.~all the data in the node belongs to the same class, and the recursion stops.
This is outlined in Lines 1 and 2 of Algorithm \ref{alg:build tree}.
Lines 4 to 7 of the algorithm generate $R$ candidate splitters at random, given $D$, $\Delta$ and $\mathrm{T}$, using Algorithm \ref{alg:generate splitters}.
Algorithm \ref{alg:generate splitters} samples $\delta$ from $\Delta$; $\tau$ from $\mathtt{T}$; and one exemplar per class from the dataset $D$ to form the set $E$.
The splitters are evaluated using Gini scores on Line 8 and the splitter with the highest Gini score is chosen o represent that node. 
After that, Lines 11 to 14 build the branches of the tree using each of the exemplars in the chosen splitter.
The tree construction process is complete once all the branches are built. 

\begin{algorithm2e}[t!]
  \caption{build\_tree($D$, $\Delta$, $\mathrm{T}$, $R$)}
  \label{alg:build tree}
  
  \DontPrintSemicolon
  \SetKwFunction{buildtree}{build\_tree}
  \SetKwFunction{ispure}{is\_pure}
  \SetKwFunction{createleaf}{create\_leaf}
  \SetKwFunction{createnode}{create\_node}
  \SetKwFunction{gensplitter}{gen\_candidate\_splitter}
  \SetKwFunction{argmax}{argmax}
  \SetKwFunction{argmin}{argmin}
  \SetKwFunction{Gini}{Gini}
  \SetKwFunction{return}{return}
  
  \KwIn{$D$: a time series dataset}
  \KwIn{$\Delta$: a set of parameterized similarity measures}
  \KwIn{$\mathrm{T}$: a set of time series transforms}
  \KwIn{$R$: number of candidate splits to consider at each node}
  \KwOut{$T$: a Proximity Tree}
  
  \BlankLine
  \If{\ispure$(D)$}{
  \return \createleaf$(D)$
  }

  \BlankLine
  \tcp{create tree, to be returned, represented as its root node}
  $T\gets \createnode()$ \;

  \BlankLine
  \tcp{Creating $R$ candidate splitters}
  $\mathcal{R}\gets \emptyset$ \;
  \For{$i=1$ \KwTo $R$}{
    \tcp{generate random splitter}
    $r\gets\gensplitter(D,\Delta,\mathrm{T})$\;
    Add splitter $r$ to $\mathcal{R}$\;
  }
  
  \BlankLine
  \tcp{select best splitter using measure $\delta^*$, transform $\tau^*$ and exemplars $E^*$}
  $(\delta^*,\tau^*, E^*)\gets \argmax_{r\in\mathcal{R}}\Gini(r)$\;

  \BlankLine
  \tcp{retain measure and transform for root node of $T$}
  $T_{(\delta,\tau)}\gets(\delta^*,\tau^*)$\;
  $T_{B}\gets\emptyset$\tcp{$T_{B}$ will store the branches under root node of $T$}
  \ForEach{\text{exemplar} $e\in E^*$}{
    \tcp{$D^*_{e}$ is the subset of $D$ closest to $e$ based on $\delta^*$ and $\tau^*$}
    $D^*_{e}\gets\{d\in D\|\argmin_{e^`\in E^*}\delta^*(d,e^`,\tau^*)=e\}$\;
    \tcp{build sub-tree for that branch}
    $r\gets\buildtree(D^*_e,\Delta,\mathrm{T},R)$\;
    \tcp{a branch is a pair (exemplar, sub-tree)}
    Add branch $(e,t)$ to $T_B$\;
  }

  \return $T$
\end{algorithm2e}

\begin{algorithm2e}[t!]
  \caption{gen\_candidate\_splitter($D$, $\Delta$, $\mathrm{T}$)}
  \label{alg:generate splitters}
  
  \DontPrintSemicolon
  \SetKwFunction{class}{class}
  \SetKwFunction{return}{return}
  
  \KwIn{$D$: a time series dataset}
  \KwIn{$\Delta$: a set of parameterized similarity measures}
  \KwIn{$\mathrm{T}$: a set of time series transforms}
  \KwOut{$(\delta, \tau, E)$: a parameterized similarity measure, transform and a set of exemplars}
  
  \BlankLine
  \tcp{sample a parameterized measure $\delta$ uniformly at random from $\Delta$}
  $\delta\xleftarrow{\sim}\Delta$ \;

  \BlankLine
  \tcp{sample a transform $\tau$ uniformly at random from $\mathrm{T}$}
  $\tau\xleftarrow{\sim}\mathrm{T}$ \;
  
  \BlankLine
  \tcp{select one exemplar per class to constitute the set $E$}
  $E\gets\emptyset$ \;
  
  \BlankLine
  \ForEach{\text{class} $c$ \text{present} $\in D$}{
    $D_{c}\gets\{ d\in D\|d.\class=c\}$ \tcp{$D_c$ is the data for class $c$} 
    $e\xleftarrow{\sim}D_c$ \tcp{sample an exemplar $e$ uniformly at random from $D_c$}
    Add $e$ to $E$
  }
  
  \return $(\delta,\tau,E)$
\end{algorithm2e}

\subsection{Transforms}
Models that learn from multiple representations has proven to be effective and accurate for TSC. 
Three of the leading algorithms, HIVE-COTE 2.0 \citep{middlehurst2021hive}, MultiRocket \citep{tan2022multirocket}, TS-CHIEF \citep{shifaz2020ts} and InceptionTime \citep{ismail2020inceptiontime} all work with multiple representations. 
HIVE-COTE 2.0 and TS-CHIEF are ensembles that leverage different time series representations and transformations including the first derivative, Shapelets, Dictionary and Intervals.
MultiRocket uses two representations, the raw and first order difference of the time series. 

It is common to transform the raw time series from their original time representation into their derivatives.
The first order derivative was first applied to $\DTW$, creating Derivative $\DTW$ ($\DDTW$) \citep{keogh2001derivative}.
The initial aim was to reduce pathological warping, that can be caused by $\DTW$, by aligning two time series in their first order derivative space instead of the raw time space.
This has subsequently been extended to other $\DTW$ variants, including $\WDTW$, as used in EE and PF 1.0 \citep{lines2015time,lucas2019proximity}.  

Other than derivatives, it is also common to transform the time series into a frequency representation, a spectrogram, using Fourier, Wavelet or other spectral transforms.
Frequency representations have been widely used in many applications such as signal processing and speech recognition. 
Besides, spectrograms have been shown to be useful for many TSC applications and used by some SoTA TSC algorithms such as HIVE-COTE and DrCIF \citep{middlehurst2021hive}.
Despite its benefit in other SoTA algorithms, spectrograms are not suitable for use with elastic similarity measures such as those used by PF.
This is because spectrograms represent the time series by ordered frequencies and it is unintuitive to align one frequency to another.
 
Therefore, we only use the first order derivative as a transform for PF~2.0.
The first order derivative is applicable to all three core similarity measures used in PF 2.0.
Then the choice of using either the first order derivative or the raw representation for a particular similarity measure is selected randomly per splitter at each node of the tree in PF 2.0, as shown in Algorithm \ref{alg:generate splitters}, where $\mathtt{T}=\{\text{raw}, \text{first derivative}\}$.
Given a time series $S=\{s_1,s_2,...,s_L\}$, we use the equation of \cite{keogh2001derivative} to calculate the first order derivative, $S'$, described in Equation \ref{eq:first derivative}.

\begin{subequations}
\begin{align}
    S'&= \frac{(s_t-s_{t-1})+(s_{t+1}-s_{t-1})/2}{2} : \forall t \in \{2,...,L-1\} \\
    s'_1 &=s'_2 \\
    s'_L &=s'_{L-1}
\end{align}
\label{eq:first derivative}
\end{subequations}

\subsection{Similarity measure and cost function parameterisation}
\label{sec:similarity}
The process of sampling a similarity measure in Algorithm \ref{alg:generate splitters} for each candidate splitter at each node can be split into three parts.
First, a similarity measure is chosen at random from the three core similarity measures, Amerced Dynamic Time Warping ($\ADTW$), Constrained Dynamic Time Warping ($\cDTW$) and Longest Common Subsequence ($\LCSS$).
This set is reduced from the initial eight core similarity measures in the original PF \citep{lucas2019proximity}.
We will show later in our Section \ref{sec:similarity selection} that the inclusion of the significantly more accurate $\ADTW$ renders many of the other similarity measures in the original set redundant. 
Besides, it has also been shown that carefully reducing the set of similarity measures in EE does not harm the overall performance  \citep{lines2022ts}.
Similar to the original PF, randomising the choice of similarity measure and its parameterization introduces variability between each tree, which increases the generalising power of the algorithm.

Second, a cost function exponent is chosen uniformly for the selected similarity measure, $\ADTW$ and $\cDTW$.
It is not intuitive to tune the cost function for $\LCSS$ as it is a measure that depends on a pre-defined threshold, $\epsilon$, where $\epsilon$ can be tuned to mimic the behaviour of tuning the cost function. 
Parameterising the cost function increases the search space for an optimal similarity measure.
The generic formula for a cost function between two points $(a,b)$ is defined as $\lambda_\gamma(a,b)=\|a-b\|^\gamma$, where $\gamma$ is the order of the cost function \citep{herrmann2023parameterizing}.
Typical values for $\gamma$ are 1 and 2, corresponding to the absolute and squared difference.
\citet{herrmann2023parameterizing} showed that learning $\gamma$ from the set $\gamma\in\Gamma=\{1/2, 1/1.5, 1, 1.5, 2\}$ improves the accuracy of $\NN$ based classifiers when used with $\DTW$-like measures. 

However, \cite{herrmann2023parameterizing} showed that computing $\DTW$ and $\ADTW$ with $\gamma=1/1.5$ and $\gamma=1.5$ takes about 7 times longer than $\gamma$ values $1/2$, $1$ and $2$. This is due to the efficient specialised implementations of $x^{1/2}$ (\texttt{sqrt}) and $|x|$ (\texttt{abs}), and $x^2$ (through multiplication of $x\times x$) on modern computing hardwares.
Therefore we sample a cost function from the set $\gamma\in\Gamma=\{1/2, 1, 2\}$ in PF 2.0 to minimise the computational overhead of calculating the exponents. 
We show in our experiments that using a smaller set does not significantly reduce the accuracy.

The final part of the similarity measure sampling process is to sample a parameter for the respective similarity measure.
The parameter sampling process for both $\cDTW$ and $\LCSS$ is the same as the original PF and we leave the exploration of a better parameter search space for future work.
The warping window parameter of $\cDTW$ is sampled uniformly at random in $[0,\frac{l+1}{4}]$, where $l$ is the length of the series.

$\LCSS$ has two parameters. 
The first parameter is a similarity threshold value, $\epsilon$, that is sampled uniformly at random from $[\frac{\sigma}{5},\sigma]$, where $\sigma$ is the standard deviation of the whole training dataset.
The second parameter is a warping window parameter that is sampled in the same way as $\cDTW$.
Recall that the $\ADTW$ penalty parameter $\omega$ is calculated from $\omega=\omega'\times r$, where $\omega'$ is the maximum penalty. 
The maximum penalty is then set to the average \emph{direct alignment} of pairs of series sampled randomly from the training dataset.

We~follow the $\ADTW$ paper for its parameterization process. We first randomly sample 4,000 pairs of series and calculate $\omega^\prime$, the average Minkowski distance between the series in each pair using the current $\gamma$. This is a sufficiently large penalty to prevent warping on most distance calculations.
Then, $\omega=(\frac{i}{100})^5\times\omega^\prime$ is set by uniformly sampling $i$ from 1 to 100. 
Note that it is possible to calculate $\omega^\prime$ separately for each node, but our experiments in Section \ref{sec:similarity selection} show that global calculation of $\omega^\prime$ gives better results on the UCR archive.
It is also important to note that $\omega^\prime$ depends on the transform and cost function used, thus it is important to first select the transform and cost function.

\subsection{Classifying with a Proximity Forest}
The classification process of a single Proximity Tree is identical to that of PF 1.0 \citep{lucas2019proximity}.
A query time series $s$ traverses down the tree from the root node to a leaf. 
At each internal node it uses the node's similarity measure, $\delta^*$, transform $\tau^*$, and exemplars $E^*$. 
It passes down the branch of the exemplar $t^*$ to which it is the nearest, $t^*=\argmin_{t\in E^*}\delta(\tau(s),\tau(t))$.
When $s$ reaches a leaf, it is  assigned the class label represented by the leaf.
Recall that the tree construction process stops when a leaf is pure, i.e.~all the instances in that leaf have the same class label.   

This process is repeated for each tree in the forest. 
A Proximity Forest then uses majority voting between its constituent Proximity Trees for the final classification.

\section{Experiments}
\label{sec:experiments}
This section describes the experiments conducted to design and assess our new Proximity Forest, PF 2.0.
We conduct a series of experiments to analyse within the proximity forest framework the relative performance of each of the original similarity measures, the new similarity measure $\ADTW$ and cost function tuning. We then experiment on combinations of these to select the small set of splitters employed in PF 2.0.
Having settled on the set of splitters, we benchmark PF 2.0's classification accuracy on the UCR time series archive.
Finally we conclude the experiments with a run-time analysis by comparing the training and inference time of PF 2.0 with PF 1.0 on the UCR datasets. 

PF 2.0 is implemented in C++ programming language.
Our code\footnote{\url{https://github.com/MonashTS/tempo}} and results\footnote{\url{https://github.com/MonashTS/ProximityForest2.0}} have been made publicly available in the accompanying websites. 
All of our experiments were conducted on a cluster with AMD EPYC EPYC-Rome 2.2Ghz Processor, 32 cores and 64 GB memory.

\subsection{Finding a better Proximity Forest}
\label{sec:ablation}
In this section, we explore the improvements on PF 1.0 that led us to PF 2.0. 
The choices explored include 
(A) tuning the cost function for the $\DTW$ family measures; 
(B) using first order derivatives transform for all measures; and
(C) the selection of measures

There is a risk that algorithms may overfit the UCR datasets, because most of the datasets are very small. 
In order to reduce overfitting on the whole UCR archive, and improving the generalisation power of PF 2.0 on larger datasets, 
the design choices are fine-tuned through stratified 5-fold cross validation.
The cross-validation datasets are created by first mixing and shuffling the default train and test sets.
Then the shuffled set is split into 5 stratified folds.   
This ensures that the training dataset is larger than in the original train/test splits, always containing 80$\%$ of the whole dataset, allowing us to better test the performance of PF 2.0 on larger datasets.

\subsubsection{Tuning the cost function and first order derivative transform}
PF 2.0 benefits greatly from tuning the cost function and from using the first order derivative transform. 
The addition of these two hyper-parameters increases the search space for the optimal similarity measure at each node, thus allowing us to later reduce the set of similarity measures used in PF 2.0 (Section \ref{sec:similarity selection}). 
In this section, we study the effect of adding first order derivative transform and cost function to PF 2.0. 
We perform the experiment on 112 UCR datasets, where they do not have missing values, variable length and have more than 1 training example per class.

\begin{figure}
    \centering
    \includegraphics[width=\linewidth]{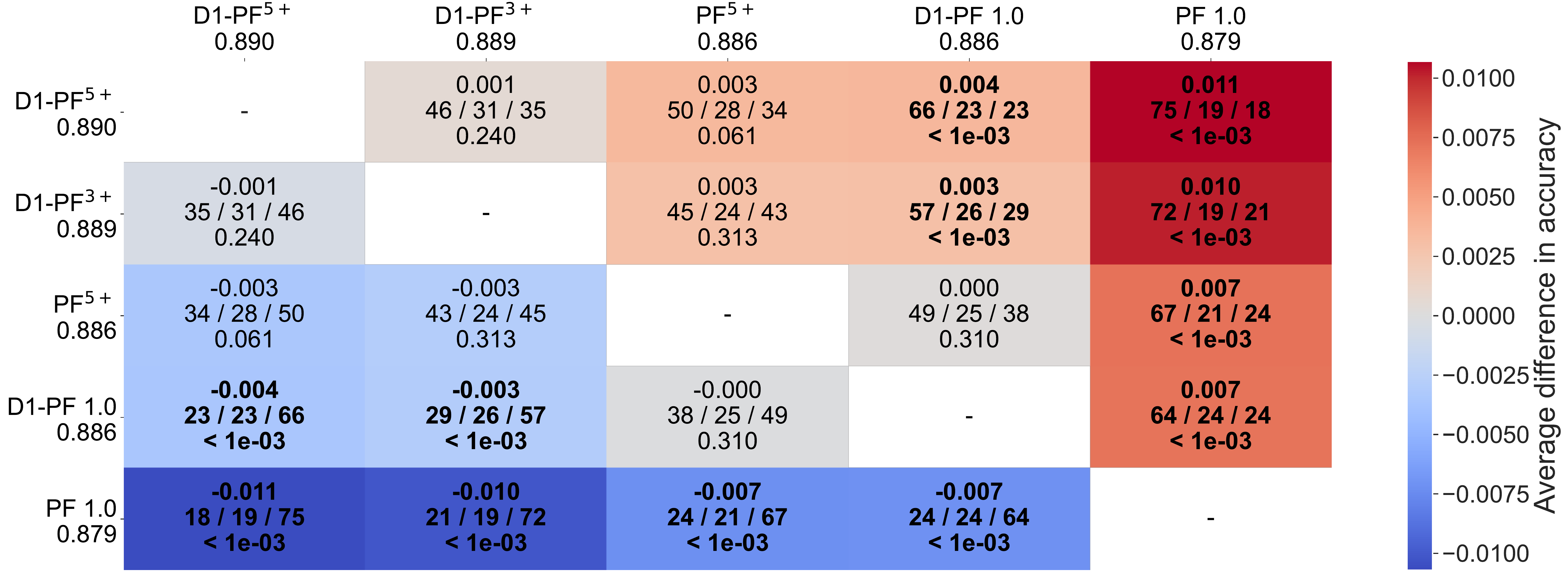}
    \caption{Pairwise statistical significance and comparison of the different PF variants.
    The methods are ranked by their accuracy on the 5-fold cross validation result of 112 UCR datasets. 
    The values in bold indicate that the two methods are significantly different under significance level $\alpha=0.05$. 
    The color represents the scale of the average difference in accuracy.}
    \label{fig:cost function tuning}
\end{figure}

Figure \ref{fig:cost function tuning} shows a multiple comparison matrix (MCM) comparing four different PF 1.0 variants---
\begin{itemize}
\item D1-PF: PF 1.0 with first order derivative transform applied to all of the eight core measures
\item PF$^{5+}$: the PF 1.0 variant proposed in \citep{herrmann2023parameterizing} with 5 cost function exponents for all the $\DTW$ variants, $\Gamma_5=\{1/2,1/1.5,1,1.5,2\}$
\item D1-PF$^{5+}$: a variant of PF$^{5+}$ by adding first order derivative transform to all of the eight core measures in PF 1.0, rather than just $\DTW$, $\cDTW$ and $\WDTW$
\item D1-PF$^{3+}$: using the smaller set of 3 cost function exponents employed by PF 2.0, $\Gamma_3=\{1/2,1,2\}$.
\end{itemize}

The methods in this matrix are ordered on the average accuracy of the method across the set of datasets.
The average accuracy is indicated below each method in the figure. 
This approach preserves the relative ordering of the methods in any comparison conducted on the same set of tasks.
Each cell in the matrix (Figure \ref{fig:cost function tuning}) contains statistics relating to a pairwise comparison between the methods on the left with the methods at the top of the column.
There are three statistics in each cell of the figure.
The~first is the average difference in accuracy between PF 2.0 and the other methods. 
The~second is the number of wins/draws/losses against the top method.
The~final row shows the p-value of a two-sided Wilcoxon signed rank significance test without multiple testing correction correction. 
We do not apply a multiple testing correction because such corrections 
1) allow a researcher to make their algorithm appear not significantly different to another simply by increasing the number of algorithms against which it is compared; and 
2) result in two algorithms being assessed differently in different studies using exactly the same outcomes.
The values in bold indicate that the two methods are significantly different at a significance level of $\alpha=0.05$.
The color in the figure represent the scale of the average difference in accuracy. 

As expected, all these PF variants are significantly more accurate across the benchmark than PF 1.0. None of the variants that use cost function tuning is significantly different to any other. All cost function tuned variants are significantly better than the variant that only extends the first derivative to all measures.
Despite being ranked the first, D1-PF$^{5+}$ is not significantly better  than D1-PF$^{3+}$ with only 3 exponents.
Due to inefficiencies in computing exponents that are not powers of 2, the set $\Gamma_3$ is significantly faster to compute than the set $\Gamma_5$ that contains the exponents 1/1.5 and 1.5, as indicated by Figure 11 in \citep{herrmann2023parameterizing}.
In consequence, we settle on the reduced cost function exponent set, $\Gamma_3=\{1/2,1,2\}$, as providing a good trade of between accuracy and speed.
This set of exponents will be used for the remaining experiments.

\subsubsection{Similarity measure selection}
\label{sec:similarity selection}
Given the superior performance of D1-PF$^{3+}$, building a D1-PF$^{3+}$ proximity forest with the core 8 similarity measures, results in an effective set of 32 similarity measures (the number of measures is doubled by first order derivative and tripled by tuning cost function applied to 4 applicable measures). 
The majority of these measures have $O(L^2)$ complexity as they do not have a warping window to reduce the computation time.
This results in D1-PF$^{3+}$ being extremely slow to compute.
In this section, we study the interactions between each similarity measures in PF with the aim of maintaining or improving the accuracy of D1-PF$^{3+}$ while being significantly faster. 

It is infeasible to experiment on all possible combinations from 2 to 9 similarity measures. 
Hence, we adopted a forward selection process by sequentially experimenting with building PF using only 1, 2, 3 and 4 similarity measures, and dropping the less effective measures as we progress to the next step.
Our results are validated using the same 5-fold cross validation approach used previously.
Instead of the 112 UCR datasets used in the previous experiments, for which is costly to run 5-fold cross validation, we follow a similar approach to that used in \cite{dempster2020rocket,dempster2021minirocket,tan2022multirocket} and work on a set of 53 `development' datasets instead.
The 53 `development' datasets are sampled randomly from the 112 UCR datasets.
This approach also helps to reduce overfitting on the entire UCR datasets.

\begin{figure}
    \centering
    \includegraphics[width=\linewidth]{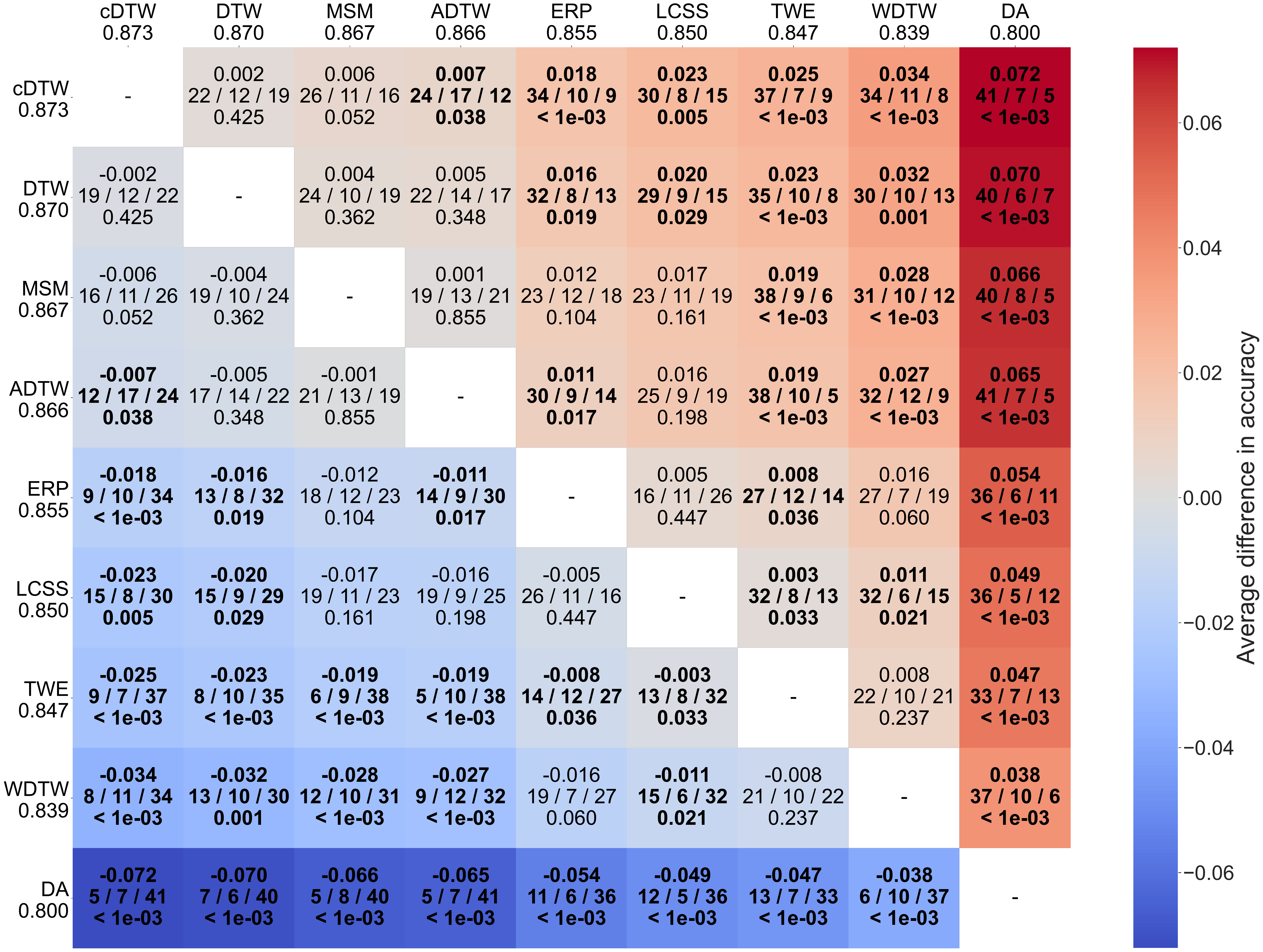}
    \caption{Pairwise statistical significance and comparison of the different D1-PF$^{3+}$ variants with 1 similarity measure.
    The methods are ranked by their accuracy on the 5-fold cross validation result of 53 UCR datasets. 
    The values in bold indicate that the two methods are significantly different under significance level $\alpha=0.05$. 
    The color represents the scale of the average difference in accuracy.}
    \label{fig:pf 1 measure}
\end{figure}

We first start by experimenting with using only a single measure in D1-PF$^{3+}$.
Each tree is built by only sampling the measure's parameters, cost function exponents (where applicable) and the choice of using the raw series or its first derivative transform.
We experimented with 9 similarity measures, including $\ADTW$ and the 8 core measures from PF 1.0.
Recall that the $\omega$ parameter for $\ADTW$ is calculated using the average \emph{direct alignment} between random pairs of time series.
In this experiment, the random pairs are sampled at each node, unlike the default setting in PF 2.0 where the pairs are sampled per tree.
We~will compare the effect of these two sampling approaches later.
Figure \ref{fig:pf 1 measure} shows the pairwise comparison of the different D1-PF$^{3+}$ variants with 1 similarity measures. 
Interestingly, the result shows that $\cDTW$ and $\DTW$ with the full window outperform all the other similarity measures, including $\ADTW$.
This shows that $\cDTW$ and $\DTW$ are robust measures for TSC.
For the remaining experiments and to maintain the feasibility of running the experiments, we keep the top 4 similarity measures ($\cDTW$, $\ADTW$, $\LCSS$, $\ERP$) and discard the rest.
$\DTW$ and $\MSM$ are discarded because they are similar to $\cDTW$ and $\ADTW$ respectively, but less accurate and slower to compute.
$\DTW$ is slower than $\cDTW$ because it does not use any warping window, while $\MSM$ is slower than $\ADTW$ due to the simpler cost function in $\ADTW$.

\begin{figure}
    \centering
    \includegraphics[width=\linewidth]{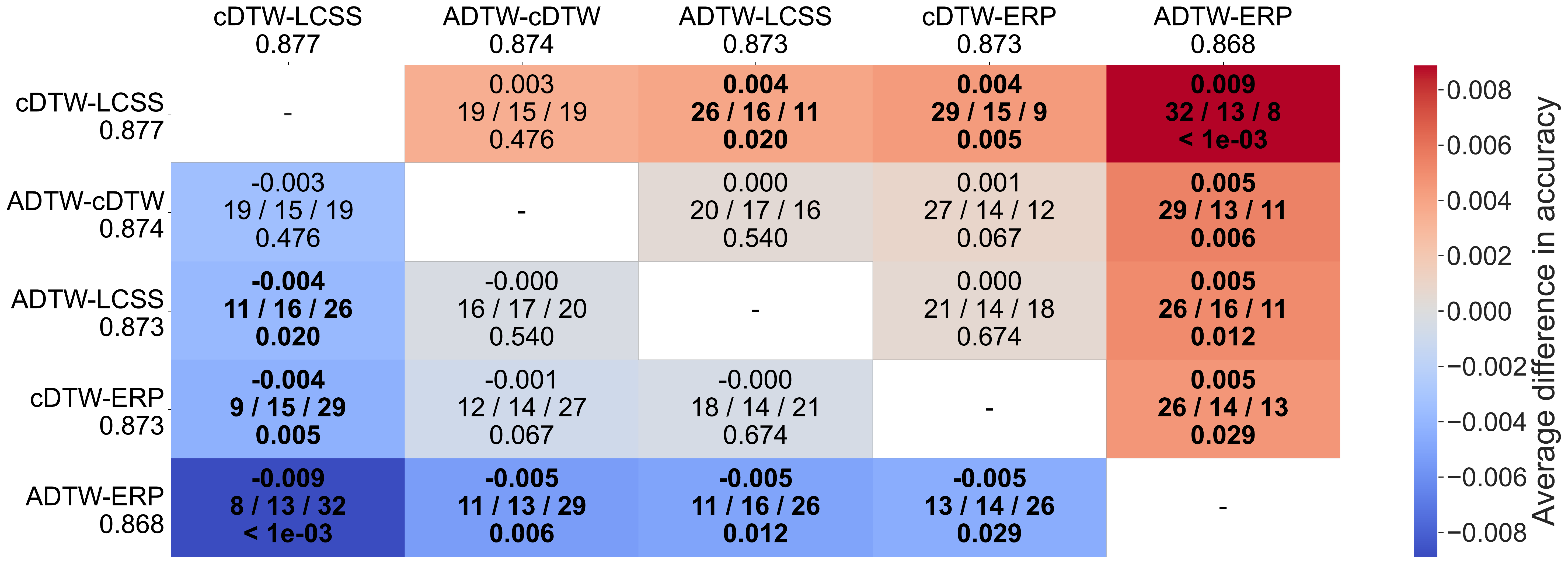}
    \caption{Pairwise statistical significance and comparison of the different D1-PF$^{3+}$ variants with 2 similarity measure.
    The methods are ranked by their accuracy on the 5-fold cross validation result of 53 UCR datasets. 
    The values in bold indicate that the two methods are significantly different under significance level $\alpha=0.05$. 
    The color represents the scale of the average difference in accuracy.}
    \label{fig:pf 2 measure}
\end{figure}

Next, we proceed with using two measures in D1-PF$^{3+}$ with the combination of the remaining similarity measures from the 1-measure experiment.
In this experiment, we only pair the top two measures $\cDTW$ and $\ADTW$ with all the other measures.
This results in 5 pairs of similarity measures as shown in the comparison heatmap in Figure \ref{fig:pf 2 measure}.  
Figure \ref{fig:pf 2 measure} shows that the top two performing pairs $\cDTW-\LCSS$ and $\ADTW-\cDTW$ outperform the other pairs.
The result also shows that adding $\ERP$ to the set reduces the performance of of $\cDTW$ and $\ADTW$ significantly.
Hence we decided to drop $\ERP$. 

With only 3 similarity measures remaining, we now shift our attention to $\ADTW$ by studying the two approaches of sampling random time series pairs for $\omega$ calculation in a proximity tree. 
The first approach is to sample the pairs at the node level, as what have been done in the previous experiments.
The second approach is to sample the pairs at the root level (per tree), which is denoted as $\ADTW_{S1}$ in this experiment.
We compare the two approaches by repeating the experiment with 1, 2 and 3 measures, as well as 4 measures including $\ERP$.

\begin{figure}
    \centering
    \includegraphics[width=\linewidth]{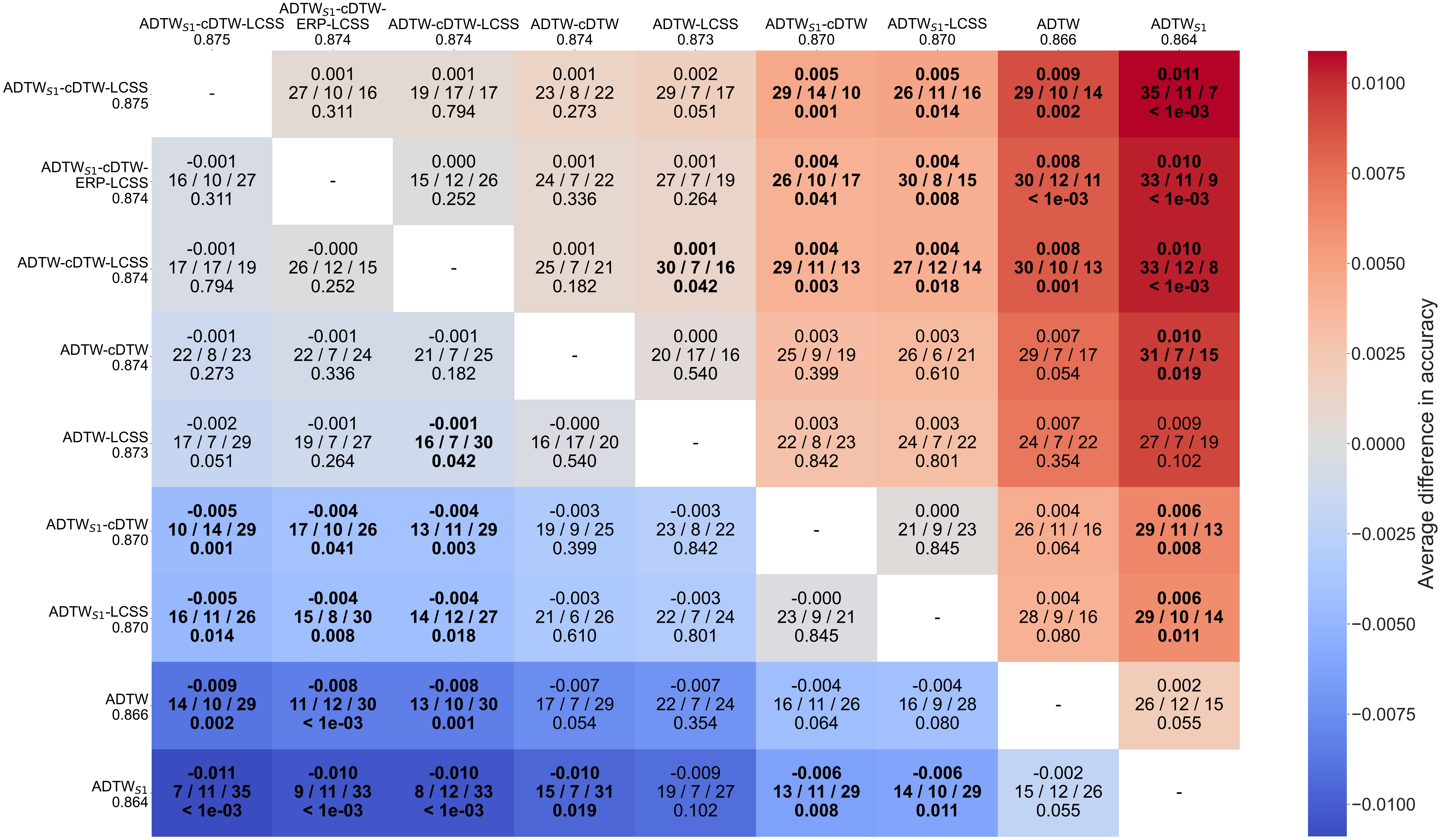}
    \caption{Pairwise statistical significance and comparison of the different D1-PF$^{3+}$ variants with 4 similarity measure.
    The methods are ranked by their accuracy on the 5-fold cross validation result of 53 UCR datasets. 
    The values in bold indicate that the two methods are significantly different under significance level $\alpha=0.05$. 
    The color represents the scale of the average difference in accuracy.}
    \label{fig:pf 4 measure}
\end{figure}

Figure \ref{fig:pf 4 measure}%
shows the pairwise comparison results of the aforementioned variants.
$\ADTW$ when paired with either $\cDTW$ or $\LCSS$ works slightly better with the first sampling approach, with their $\ADTW_{S1}$ counterparts ranked lower, albeit not significantly so. In contrast, sampling at the root is slightly higher ranked, but not significantly so, when paired with both $\cDTW$ and $\LCSS$.
Hence, we use the root node sampling approach as it is computationally cheaper. 
The pairwise result also shows that the variant with 4 measures is ranked lower than the 3 measures.
In consequence, we adopt the top performing model, D1-PF$^{3+}$ with $\ADTW_{S1}$, $\cDTW$ and $\LCSS$ as PF 2.0.

So far we have recommended a set of similarity measures that works well in general on the UCR benchmarking archive. 
As different measures work differently on different types of datasets, it is possible to fine tune the set of distances for each individual dataset.
We believe this enhancement will significantly improve the accuracy of PF 2.0 and consider this as future work. 





\subsection{Run-time analysis}
\begin{figure}
    \centering
    \begin{subfigure}{0.49\textwidth}
        \includegraphics[width=\textwidth]{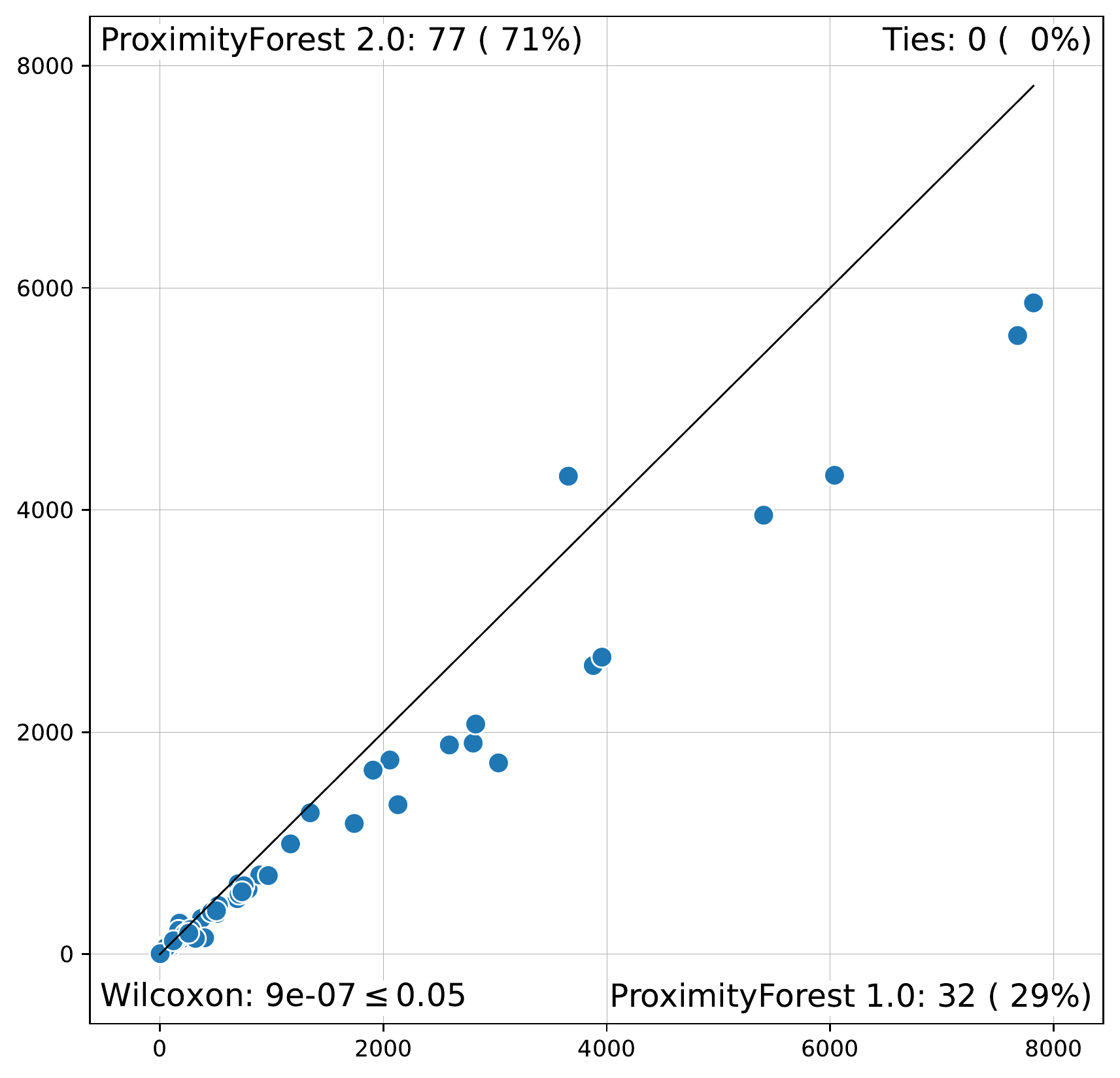}
        \caption{}
        \label{fig:train time}
    \end{subfigure}
    \begin{subfigure}{0.49\textwidth}
        \includegraphics[width=\textwidth]{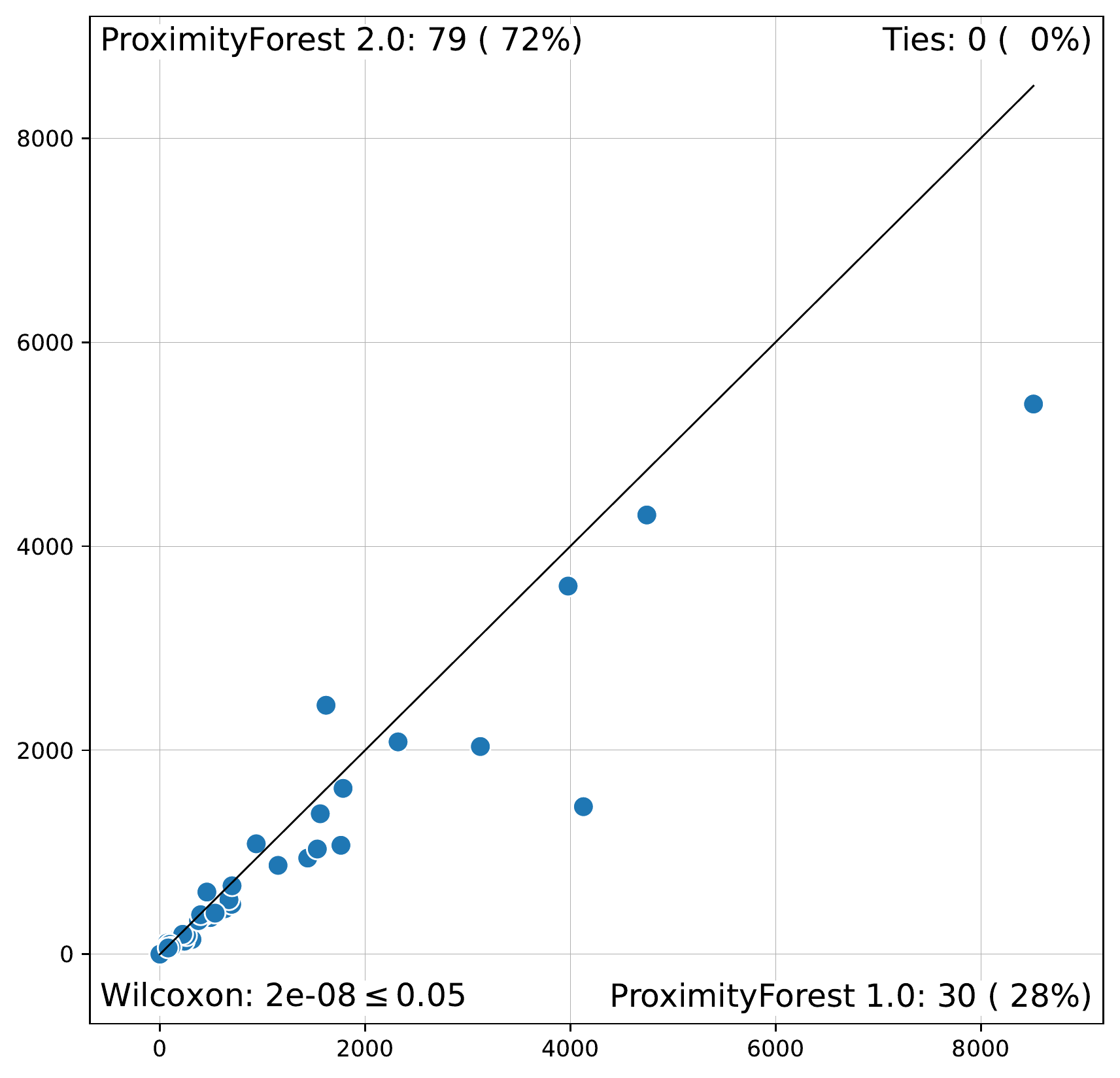}
        \caption{}
        \label{fig:test time}
    \end{subfigure}
    
    \caption{(a) Train and (b) test time comparison between PF 1.0 and PF 2.0. Each point is a dataset and points below the diagonal indicates that the method on the y-axis is faster and vice versa.}
    \label{fig:train and test time}
\end{figure}

In this section, we compare the run-time of PF 2.0 with the original PF that uses 11 similarity measures.
PF 2.0 has three core similarity measures. 
Each of them uses first order derivative transform and two of them use a set of three cost function exponents.
This makes a total of 14 effective similarity measures in PF 2.0. 
Figure \ref{fig:train and test time} compares the train time (Figure \ref{fig:train time}) and test time (Figure \ref{fig:test time}) using AMD EPYC EPYC-Rome 2.2Ghz Processor with 1 core and 64 GB memory on 109 UCR datasets -- a subset of the 112 datasets without 3 large and long datasets\footnote{\texttt{HandOutlines} with 1000 training examples and 2709 in length; \texttt{NonInvasiveFetalECGThorax1} and \texttt{NonInvasiveFetalECGThorax2} both with 1800 training examples and 750 in length}.
Surprisingly, the plots show that PF 2.0 with more similarity measures is faster than PF 1.0, especially on larger and longer datasets, showing the efficiency of PF 2.0. 
The total train and test time for PF 2.0 is 15 and 10 hours respectively while it is 20 and 13 hours for PF 1.0.
This is likely to be because most of the similarity measures in PF 2.0 are fast to compute compared to PF 1.0. 
For instance, $\cDTW$ uses a warping window to speed up the computation and $\ADTW$ has a cheap cost function.   

While the training and testing time on a single CPU has improved, we also ran the same timing experiment using 31 CPU cores.
This gives the total train and test time for PF 2.0 to be approximately 42 minutes and 1 hour respectively.
The improvement in timing is a result of the efficient implementation of PF 2.0 in C++ and leveraging EAP \citep{herrmann2021early}. 
The source code for PF 2.0 is now available as a C++ package library.

\subsection{Benchmarking}
In this section, we evaluate the classification accuracy of PF 2.0 and benchmark it on the UCR time series classification archive.
The results for both PF 1.0 and PF 2.0 are obtained with the default settings, $K=100$ trees and $R=5$ candidate splitters. 
We evaluate PF 2.0 on 109 datasets from the UCR archive.
This is a subset of the 112 datasets where 3 large and long datasets are dropped as they are too costly to compute by other state of the arts.
Rather than the original training and test splits, we use the standard 30 stratified resamples from the merged data that have been used by many other researchers \cite{middlehurst2021hive,bagnall2017great,tan2022multirocket} 

\begin{figure}
    \centering
    \includegraphics[width=\linewidth]{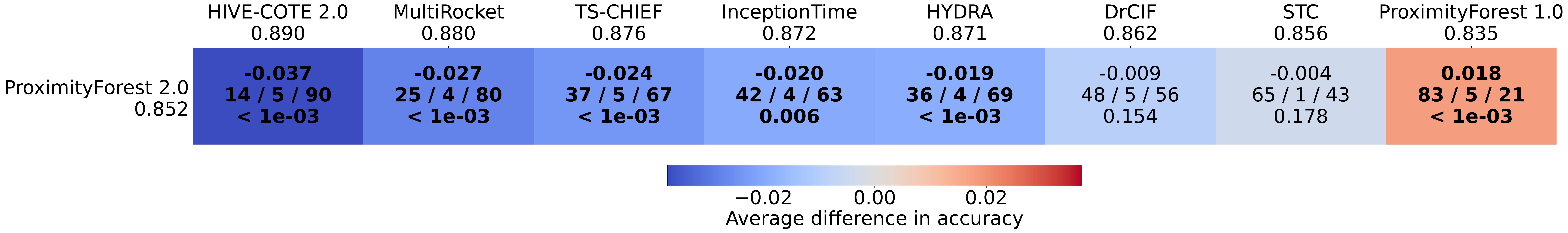}
    \caption{Pairwise comparison of PF 2.0 with key SoTA methods.
    Each cell presents the average difference in accuracy across all datasets, the win/draw/loss counts of numbers of datasets for which PF 2.0 obtains higher or lower accuracy and the p-value from a Wilcoxon signed rank test.
    The methods are ranked by their average accuracy across all 30 resamples of 109 UCR datasets, indicated by the values below each method. 
    The values in bold indicate that the two methods are significantly different under significance level $\alpha=0.05$. 
    The color represents the scale of the average difference in accuracy.} 
    \label{fig:heatline}
\end{figure}

We compare PF 2.0 with the current most accurate TSC algorithms, namely HIVE-COTE 2.0, TS-CHIEF, InceptionTime, MultiRocket, DrCIF, HYDRA, STC and PF 1.0.
These algorithms were chosen as the most accurate in their respective domains. 
The results are presented in Figure \ref{fig:heatline} in the form of a multiple comparison matrix, as seen in previous experiments.
These results indicate that the mutiple domain algorithms HIVE-COTE 2.0 and TS-CHIEF and convolutions algorithms MultiRocket, INceptionTime and HYDRA all significantly outperform PF 2.0 across the benchmark. 
PF 2.0 performs at a similar level to the leading single domain interval and dictionary techniques DrCIF and STC. 
It significantly outperforms PF 1.0.

\subsection{Comparison with similarity-based methods}
In the following, we aim to compare PF 2.0 with the following similarity-based approaches:
\begin{enumerate}
    \item $1\NN{-}\ADTW^{+}$ -- the most accurate $1\NN$ classifier, using $\ADTW$ with leave-one-out tuning of its $\omega$ warping penalty and $\gamma$ pairwise alignment cost function \citep{herrmann2023parameterizing}.
    \item Elastic Ensemble (EE) -- the previous second most accurate similarity-based approach with 11 $1\NN$ classifiers \citep{lines2015time}.
    \item TS-QUAD -- a smaller version of EE with only 4 similarity measures \citep{lines2022ts}.
    \item EE$^{+}$ -- a variant of EE where the cost function of $\DTW$-like measures ($\DTW$, $\cDTW$, $\WDTW$, $\ED$) are tuned.  
    \item EE$_{PF2}$ -- a variant of EE that uses the same set of similarity measures as PF 2.0 as well as tuning the cost function.
    \item PF 1.0 -- the first PF and the current most accurate similarity-based approach with 11 similarity measures \citep{lucas2019proximity}.
    \item PF$^{+}$ -- a variant of PF 1.0 where the cost function of $\DTW$-like measures are tuned \citep{herrmann2023parameterizing}.
\end{enumerate}

\begin{figure}
    \centering
    \includegraphics[width=\linewidth]{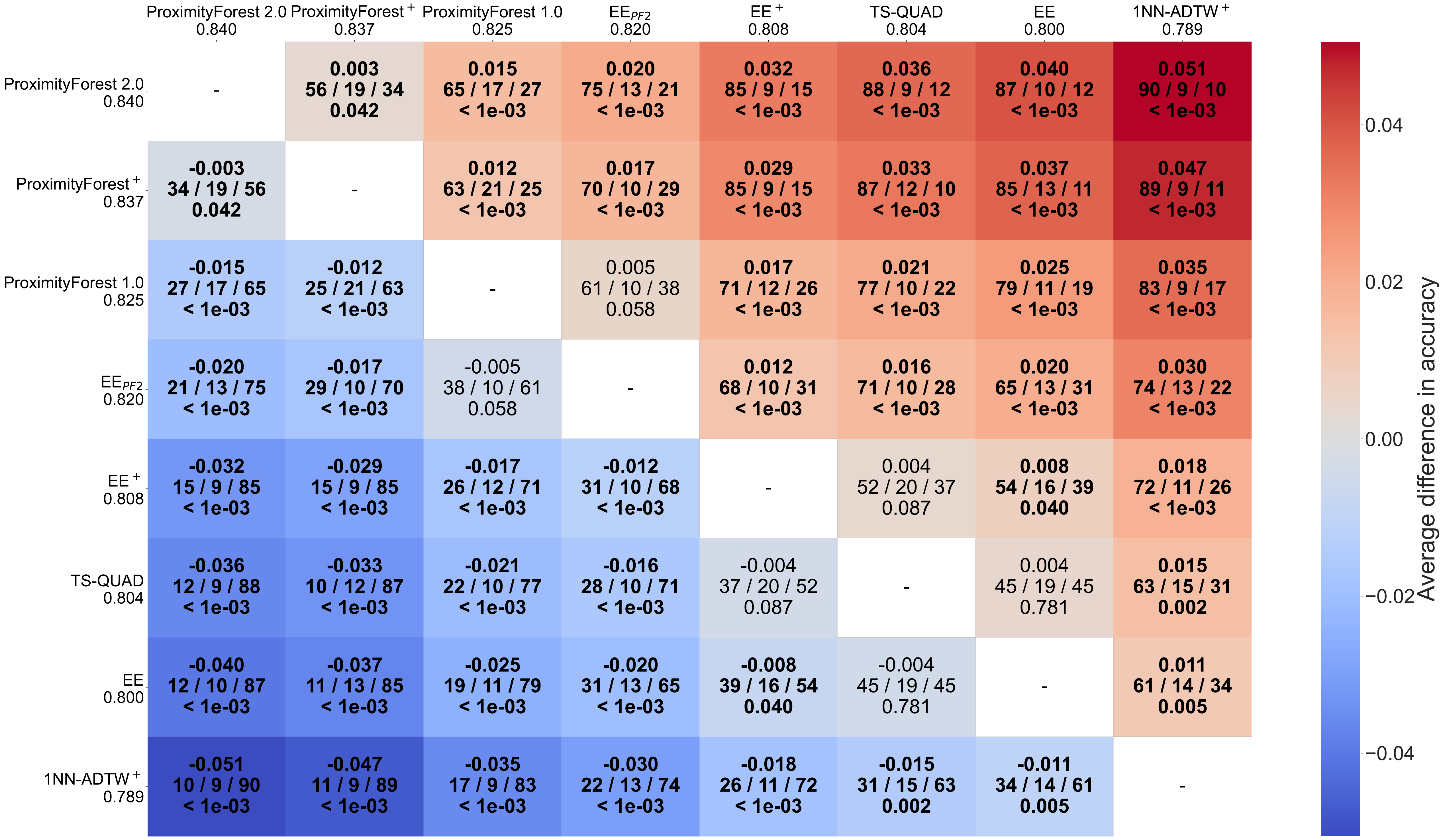}
    \caption{Pairwise statistical significance and comparison of PF 2.0 with other similarity-based approaches.
    The methods are ranked by their average accuracy on the default resample of 109 UCR datasets. 
    The values in bold indicate that the two methods are significantly different under significance level $\alpha=0.05$. 
    The color represents the scale of the average difference in accuracy.}
    \label{fig:similarity_based}
\end{figure}

Since EE is not scalable and costly to compute, the methods were compared only on the original train/test split of 109 UCR datasets. 
Figure \ref{fig:similarity_based} shows that PF 2.0 is significantly more accurate than any of the similarity-based approaches.
Both PF$^{+}$ and EE$^{+}$ show that tuning the cost function improves the classification performance compared to the original variant.
The higher ranked EE$_{PF2}$ than all other EE variants including TS-QUAD, proves that this new set of similarity measures are not only useful for PF but also useful for other similarity-based approaches such as EE.
We provide additional studies in Section \ref{sec:ablation} to further strengthen our claims here and justifying the design choices in PF 2.0 such as the use of cost function tuning, first order derivatives transform and the set of similarity measures.

\section{Conclusion}
\label{sec:conclusion}
We have developed Proximity Forest 2.0 -- a new Proximity Forest that is faster and more accurate than the original.
PF 2.0 incorporates three important recent advances in time series similarity measures (1) computationally efficient early abandoning and pruning to speedup elastic similarity computations, (2) a new elastic similarity measure, Amerced Dynamic Time Warping ($\ADTW$) and (3) cost function tuning. 
We have also implemented PF 2.0 in C++, making it more efficient and scalable than before. 
Our experimental results show that PF 2.0 is the most accurate and fastest similarity-based TSC algorithm benchmarked on the UCR archive. 
We also show that using more similarity measures in PF 2.0 could hurt performance and the set of similarity measures should be carefully selected so that it does not overfit the benchmark datasets.

PF 2.0 uses three core distance measures, $\ADTW$, $\cDTW$ and $\LCSS$, plus a variant of each that adds the first derivative transform, a total of six distance measures, four of which are also coupled with cost function tuning. We have shown the the Elastic Ensemble also benefits from using this set of measures, greatly increasing its efficiency while also improving its accuracy.  This suggests that this set of measures might be useful in further contexts. 

We believe that there is potential to fine tune the set of distances for each individual dataset, which could further improve the accuracy of PF 2.0 and plan to consider this as future work. 

\section*{Acknowledgments}
\sloppy This work was supported by the Australian Research Council award DP210100072.
The authors would like to thank Professor Eamonn Keogh and his team at the University of California Riverside
(UCR) for providing the UCR Archive.

\bibliographystyle{plainnat}
\bibliography{references}

\end{document}